%% file: main.tex
\documentclass[twocolumn]{ceurart}
\usepackage{soul}

\usepackage{graphicx} 

\usepackage{amsmath, amssymb, dsfont}
\usepackage{physics}
\usepackage{amsfonts}
\usepackage{subcaption}   
\usepackage{booktabs}
\usepackage{arydshln}
\usepackage{pifont}

\usepackage{xcolor}

\usepackage{listings}
\usepackage{multirow}

\date{March 2024}

\begin{document}

\title{Skill matching at scale: freelancer-project alignment for efficient multilingual candidate retrieval}

\author[1]{Warren Jouanneau}[%
orcid=0000-0003-4973-2416,
email=warren.jouanneau@malt.com,
]
\author[1]{Marc Palyart}[%
]
\author[1]{Emma Jouffroy}[%
]
\address[1]{Malt, 33000 Bordeaux, France}

\input{0_abstract}
\maketitle

\input{1_introduction}

\input{2_related_work}

\input{3_proposition}
\input{4_results}

\input{5_industrial}

\input{6_futur}

{\small
\bibliography{biblio}}
\end{document}

%% file: 0_abstract.tex
\begin{abstract}
Finding the perfect match between a job proposal and a set of freelancers is not an easy task to perform at scale, especially in multiple languages. In this paper, we propose a novel neural retriever architecture that tackles this problem in a multilingual setting. Our method encodes project descriptions and freelancer profiles by leveraging pre-trained multilingual language models. The latter are used as backbone for a custom transformer architecture that aims to keep the structure of the profiles and project. This model is trained with a contrastive loss on historical data. Thanks to several experiments, we show that this approach effectively captures skill matching similarity and facilitates efficient matching, outperforming traditional methods.
\end{abstract}


%% file: 1_introduction.tex
 \begin{figure*}[]
\centering
\subfloat[]{\includegraphics[width=0.45\textwidth]{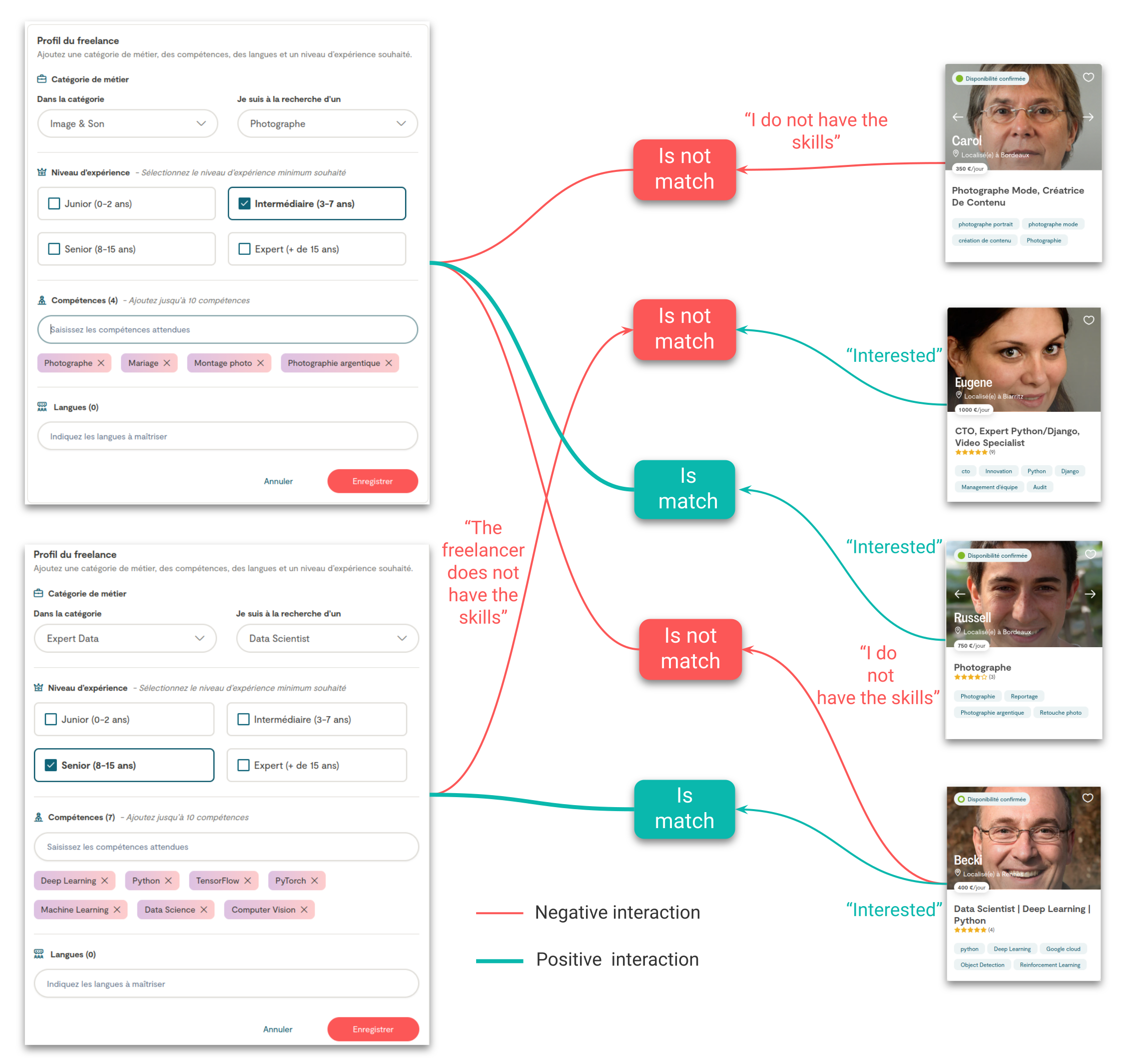}%
\label{fig:pairs_interactions}}
\hfil
\subfloat[]{\includegraphics[width=0.45\textwidth]{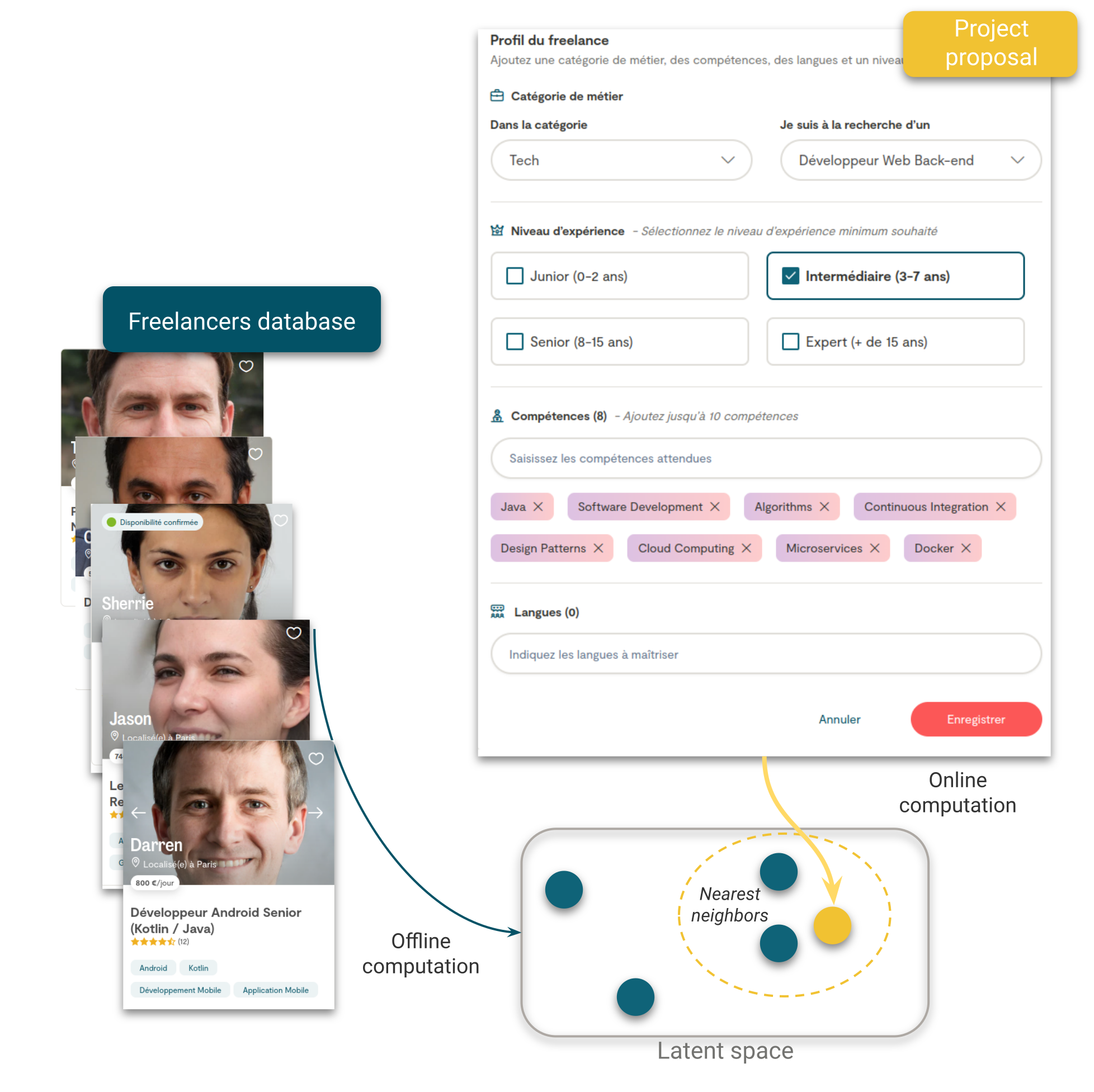}%
\label{fig:latent_projection}}
\caption{Illustration of matching interactions and latent space projection. (a) : Examples of positive and negative interactions between a project proposal and a set of freelancers. (b) : Illustration of documents projection and retrieval within the latent space.}
\label{fig:methodillustration}
\end{figure*}

\section{Introduction}
\label{sec:intro}

With more than 700,000 registered freelancers, Malt is the leading freelancing platform in Europe.
On the platform, users have the option to search for freelancers on their own or to post a project that will be handled fully by our recommender system.
Like other players in the human resources field \cite{kenthapadi2017personalized, geyik2018talent} we have been using machine learning for several years to help match projects to freelancers. Our recommender system takes as input a project description and contacts automatically a set of relevant freelancers to see if they are interested in the project. 

At the time, this system suffered from three limitations. First, we could not scale well, as each new project needed to be scored against each freelancer to build the global ranking. Second, the system only relied on partial information contained within the rich freelancer profiles. For example, job title and skills were taken into account but long text sections such as profile description or experience description were under-exploited. Finally, as Malt is active across Europe, language management was painful since skill matching model were monolingual and one had to be developed and maintained for each language supported. Plus, the cross-lingual matching was badly handled. In summary, we needed a multilingual approach that could scale and use richer information than the legacy system.

Our legacy system had just the filtering and ranking phases. To map to a traditional recommender architecture that can scale, we decided to add a retrieving phase to generate relevant candidates. Traditionally, two approaches are used \cite{Mitra2018:NeuralIR}. One based on lexical matching (bag-of-words models), and another called semantic matching based on representational learning (neural network models). However, the former suffers from a lack of semantic and context interpretation, an issue that the latter is supposed to solve.

Indeed, in our case, there are fundamental differences in the nature and form of information disclosed by freelancers and project proposals. Freelancers and companies may employ different vocabularies, potentially due to varying levels of expertise. Finally, projects typically require a subset of a freelancer's skills, resulting in less specific information being communicated. 

Therefore, our goal is to build upon existing work, such as SentenceBERT~\cite{sentence-bert} and conSultantBERT, within a multilingual setting~\cite{sentence-bert-multilingual,e5-multilingual},  to develop an efficient freelancer retrieval phase. This retrieval model should reflect our historical data, as illustrated in Fig. \ref{fig:pairs_interactions}, and enable the selection of candidates based on skill-matching similarity, as shown in Fig. \ref{fig:latent_projection}.
The proposed method relies exclusively on skills, as business-related factors such as experience or location are managed separately within our pipeline, either through the ranker or via hard filtering rules.
The retrieval approach, along with the associated experimental results, are presented in this article as follows : 
\begin{itemize}
    \item Part 2 examines prior related research and discusses their limitations within the context of our work.

    \item Part 3 provides an in-depth explanation of our approach, which consists of an architecture that leverages a multilingual backbone while effectively handling documents and their structures, as well as a training loss and configuration designed to facilitate retrieval.

    \item Part 4 outlines the experimental protocol and the models tested. In addition, new evaluation metrics are proposed.

    \item Finally, Part 5 provides insights into the results obtained after deploying the proposed model in production.
\end{itemize}

%% file: 2_related_work.tex
\section{Related work}

The proposed approach, while similar to traditional job board methods, adopts a reversed recommendation strategy: it recommends freelancers to employers rather than job posts to candidates. Despite this perspective shift we can benefit from current approaches in the human resources (HR) domain.
Indeed, in recent years, available data in HR have enabled to enhance person-job fit algorithms. By improving how candidates are matched with job roles, these advancements have made hiring processes more efficient and effective, ensuring that the right talent meets the right opportunity. Various approaches, including content-based filtering \cite{mpela2020mobile,chenni2015content,bansal2017topic}, collaborative filtering \cite{lee2017exploiting,reusens2017note,ahmed2016user}, and hybrid strategies, have emerged \cite{luo2019resumegan,bian2020learning,dave2018combined}.

Due to their principal modality, documents containing text, most of these methods utilize embeddings of input data to measure similarity, a critical aspect of evaluating a candidate's fit for a particular role. The richness of embeddings is explored in multiple studies \cite{jiang2020learning, kaya2021effectiveness, lacic2019should}, which demonstrates that integrating these features can enhance job matching accuracy.

Approaches vary on embedded textual information types and techniques used, such as Word2Vec and Doc2Vec \cite{kaya2021effectiveness, lacic2019should}, the exploration of graphical representations, \cite{yang2022modeling}, the use of deep learning techniques, such as Convolutional Neural Networks (CNNs), Recurrent Neural Networks (RNNs), and Long Short-Term Memory (LSTM) networks \cite {qin2018enhancing, ramanath2018towards}. More recently, methods based on the attention mechanism have emerged, such as PJFCANN \cite{wang2022person} or conSultantBERT~\cite{lavi2021consultantbert}. Additionally, these approaches can be adapted to handle multiple languages through distillation \cite{sentence-bert-multilingual}. To account for the inherent structure of documents, whether résumés or job proposals, all sections can be concatenated~\cite{lavi2021consultantbert}, processed at the sentence level~\cite{le2014distributed}, or processed at the paragraph level~\cite{dai2015document}. However, to our knowledge, no existing method processes each section while preserving its specificity. This results in the loss of the section's intrinsic type information.

The aforementioned models transform textual data through multiple non-linear transformations, resulting in more abstract representations. They create a framework for manipulating the representational space, referred to as the latent space. The emerging field of representation learning \cite{bengio2013representation} focuses on using deep learning to acquire meaningful representations within this latent space, improving predictions for higher-level tasks such as classification, regression or clustering.

The choice of representation learning technique depends on the input modality and is tailored to the downstream task \cite{radford_unsupervised_2016, bengio2013representation}.

Regarding retrieval process within HR applications, natural clustering representation is highly desirable for similar entities. 
Most approaches rely on contrastive representation learning \cite{le2020contrastive}. The latter type of methods aim to minimize the embedding distance for similar entities, while maximizing it for dissimilar ones. Person-job fit systems often use these approaches but face challenges such as sparse or unstructured data and large pools of freelancers and clients, leading to significant computational costs.

To improve contrastive learning, researchers have focused on enhancing settings, relational data, and loss functions. For settings, models like CrosCLR \cite{crosclr} explore intra-modality relations using matrices, which is relevant as profiles and project proposals can be seen as different modalities. In terms of relational data, the number of counterexamples significantly impacts performance. Some approaches also dynamically select hard examples at the batch level during training \cite{hermans2017defense}, called online batch mining. There are various loss functions and advancements available for training models in a contrastive manner. Entities can be presented to the models in pairs, using either a classification or regression loss~\cite{sentence-bert}, as was explored in the human resource context with conSultantBERT \cite{lavi2021consultantbert}. Alternatively, the input can be structured as triplets, consisting of one anchor entity, one relatively positive entity, and one negative entity. Initially proposed in computer vision~\cite{schroff2015facenet}, the \textit{triplet loss} has also been applied to text models, such as SentenceBERT~\cite{sentence-bert}. In the literature, using triplets instead of pairs has often been shown to yield better results. Based on this observation, several attempts have focused on incorporating multiple negatives and/or multiple positives simultaneously. One such approach is the InfoNCE loss~\cite{oord2018representation}. The latter is promising due to its automatic hard example mining. Models like CDC \cite{CDC} maximize the mutual information of future data in latent space using recurrent neural networks. This approach has been adapted for image and unsupervised patch representation and can potentially be generalized to other representation types. Moreover, the Supervised CDC model \cite{Supervised-CDC-Image} reintroduces a supervised setting, allowing generalization to any number of positive examples. However, this approach has yet to be evaluated on the human resource domain.

The following section describes our proposition of a model handling documents structures in different languages: a neural encoder based on contrastive learning using InfoNCE.

%% file: 3_proposition.tex
\section{Approach}
\label{approach}

In the following, we denote project proposals as $x_p$ for all project $p \in \mathbb{P}$, where $\mathbb{P}$ is the set of all possible projects. Similarly, freelancers' profile are denoted as $x_f$ for all freelancer $f \in \mathbb{F}$, where $\mathbb{F}$ is the set of all possible freelancers. This section aims to describe two-tower encoder models, denoted as $M_F$ and $M_P$, which encode and enable the retrieval of freelancers based on submitted projects. \\
Generic documents (\textit{i.e.} either a proposal or a profile) are denoted as $x_d$, for all projects or freelancers $d \in \{\mathbb{F} \cup \mathbb{P}\}$. Each document consists of different sections, denoted as $s_{d,l}$, for all $l \in L_{\text{section}}$ and $d \in \{\mathbb{F} \cup \mathbb{P}\}$. Here, $l$ represents a type among the $L_\text{section}$ different section types. Each section is plain text, and thus, each document is a set of section texts, denoted as:
\[ x_d = \{s_{d,l} \mid l \in L_{\text{section}}\}. \] 

The role of the models is to encode these documents into embedding vectors. Specifically, we have:
\[ M_F(x_f) = \vb*{e}_f \quad \text{and} \quad M_P(x_p) = \vb*{e}_p. \]

However, these embeddings must lie in the same representation space, where the distance between them is semantically meaningful in terms of skill matching. Projecting a project into this space allows for retrieval by proximity. To achieve this, we leverage past interactions between freelancers and projects using a contrastive loss function. \\
As illustrated in \textit{Fig.} \ref{fig:methodillustration}, for past interactions, we consider a recommended freelancer as a negative match for a project if either the company or the freelancer himself declares that the freelancer does not have the skills. Such cases are considered negative project-freelancer pairs. If the freelancer replies that they are interested in the project, then the pair is considered positive. If no feedback is provided, the pair is discarded from the dataset.\\
In the subsequent sections, we first present the architecture before describing  the loss used for training the models and obtaining a semantically meaningful representation space.

\subsection{Architecture}

\begin{figure*}
    \centering
    \includegraphics[width=\linewidth]{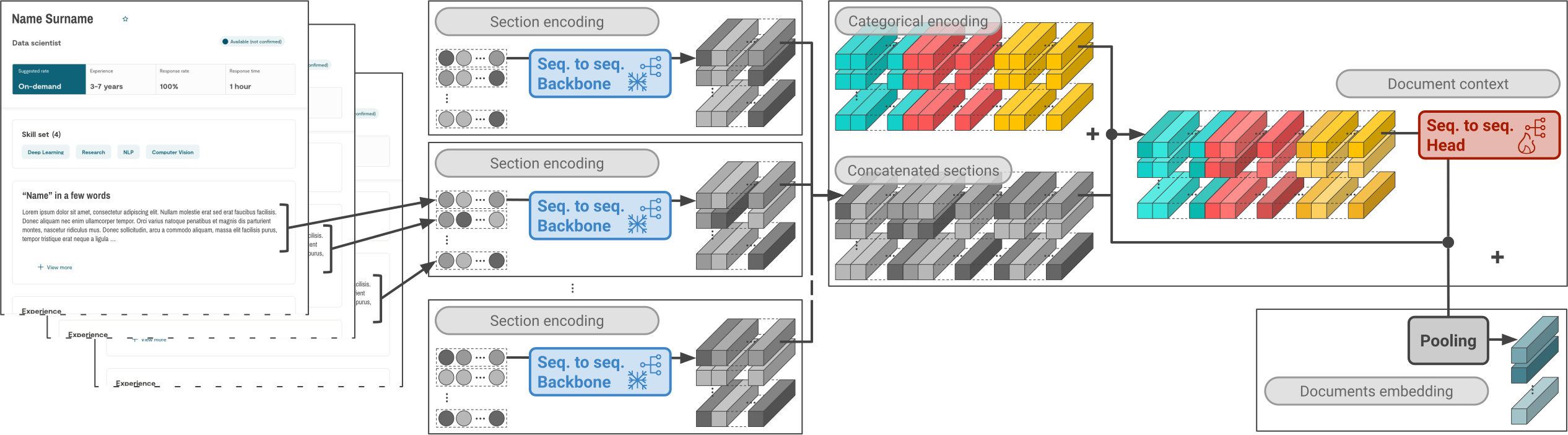}
    \caption{Illustration of the proposed architecture for the freelancer model. Similar sections across profiles are projected into a latent space (shown in the first three squares). Positional encodings and resulting embeddings are then combined to retain section type information before entering a transformer head to process document-level context. The final step involves pooling, producing the document embedding.}
    \label{fig:architecture}
\end{figure*}

For both proposals and profiles document types, the same architecture $M$ is used for both models ($M_F$, $M_P$) of the two tower approach. Therefore, the two models 
\[ M_F(x_f) = M(W_F, x_f) \quad \text{and} \quad M_P(x_p) = M(W_P, X_p), \]
depend on the trained weights $W_P$ and $W_F$ respectively. For a generic document type $x_d$, whether a profile or a proposal, the following detailed architecture $M$ is designed to process the input, resulting in its corresponding embedding vector $M(W, X_d) = \vb*{e}_d$. This embedding vector will subsequently be utilized for retrieval purposes.\\
The proposed architecture is designed to process any document type consisting of text, with the objective of leveraging its inherent structure, while keeping language alignment.
It is depicted in \textit{Fig.} \ref{fig:architecture} and can be summarized as follows : 
A pretrained multilingual backbone is used to capture local context, specifically at the section level,
while handling different languages, and is described in section \ref{section_level} (\emph{Section encoding} on \textit{Fig.} \ref{fig:architecture}). A positional encoding is added for the model to carry the indication of the section type, as explained in section \ref{positional_encoding} (\emph{Categorical encoding} on \textit{Fig.} \ref{fig:architecture}). In complement, a global context processing at the document level is introduced through a dedicated transformer head, which is detailed in section \ref{document_level} (\emph{Document context} on \textit{Fig.} \ref{fig:architecture}). Both components ensure a comprehensive understanding of local and global contexts within the document, while the use of a frozen pre-trained multilingual backbone ensure language alignment. Finally, the section \ref{emb_to_vec} explains how embedded sections are balanced for obtaining the final document representation (\emph{Documents embedding} on Fig. \ref{fig:architecture}).

\subsubsection{Leveraging pre-trained multilingual model : section-level context}
\label{section_level}

By considering a document as a sequence of sections, and each section as a sequence of tokens, it becomes possible to process these sections independently using state-of-the-art pre-trained language models. \\
To encode the sequence of tokens $t_i$ with a length of $k_l$ within a section $s_{d,l} = \{t_i \mid i=1,\dots, n \}$, we utilize a sequence-to-sequence transformer \cite{sutskever2014sequence}, which serves as the backbone of our architecture, represented as follows:
\begin{equation}
\operatorname{Backbone}(s_{d,l}) = E_{\text{token}_{d,l}} = \{\vb*{e}_{\text{token}_{d,l,i}} \mid i=1\dots k_l \}.
\end{equation}
By processing all sections using the same model, we ensure that each token embedding $\vb*{e}_{\text{token}_{d,l,i}} \in E_{\text{token}_{d,l}}$ captures context at the section level. This is achieved through the transformer model's architecture and its attention mechanism, which focuses on the relationships within the section's sequence of tokens. \\
The choice of a multilingual similarity backbone has been made considering the need for a latent space that is both organized based on semantic similarity \cite{sentence-bert}, and possesses language alignment \cite{sentence-bert-multilingual}. Such models are considered as great semantic extracting encoders but encode the full sequence of tokens as a unique embedding vector using a pooling layer. Moreover, the resulting embedding is often fed into a projection layer, which participates in the latent space organization. \\
Hence, using a model without a projection layer after pooling will allow semantic similarity and language alignment to be directly reflected in the token embeddings. In other words, the token embeddings $\vb*{e}_{\text{token}_{d,l,i}} \in E_{\text{token}_{d,l}}$ of a section $s_{d,l}$ will inherently maintain semantic alignment across different languages. Therefore, we assume that using such a model as a sequence-to-sequence backbone (\textit{i.e.} with no final projection layer), omitting the final pooling layer and freezing its weights, will result in the conservation of the semantic space organization and the alignment across languages. \\
The following sections describe the adaptation of this backbone and the specification of the whole architecture to the freelancer-project domain.

\subsubsection{Differentiating between section : positional encoding}
\label{positional_encoding}

By concatenating the encoded token sequences from each section, $E_{\text{token}_{d,l}}$, we can construct a single sequence for the entire document, resulting in $E'_{\text{token}_{d}}$. \\
However, concatenating the sequences leads to the loss of information regarding the original section from which each token is derived. While each token in the newly formed sequence retains its semantic meaning, section-level context, and positional information, it no longer carries any indication of its section type.\\
To alleviate this issue, we propose the use of categorical encoding to incorporate section type information at the token level, similar to how positional encoding provides positional information to the model. To achieve this, we utilize learned embeddings $\vb*{e}_{\text{categorical}_{l}}$ for each section type label $l \in L_{\text{section}}$, and add these learned weights to the token embeddings. This process is formalized as follows:
\begin{align}
E'_{\text{token}_{d}}  &= \bigodot_{l \in L_\text{section}} \{\vb*{e}_{\text{token}_{d,l,i}}+\vb*{e}_{\text{categorical}_{l}} \mid i = 1...k_l\} \notag\\
&= \{ \vb*{e'}_{\text{token}_{d,l,i}} \mid i=1 \dots n \},
\end{align}
where the length $n$ of the sequence is the sum of the lengths of all sections in the document, given by $n = \sum_{l \in L_\text{section}} k_l$, and $\odot$ denotes the concatenation operation.

\subsubsection{Introducing a transformer head : document-level context}
\label{document_level}

At this stage, each token represents section-level context and remains language-agnostic due to the use of the frozen backbone model. To incorporate document-level context, we process the concatenated sequence $E'_{\text{token}_{d}}$ using a BERT-like sequence-to-sequence transformer \cite{BERT} as an architecture head. This trained model head leverages the attention mechanism to enrich each token's embedding with context from tokens across different sections. \\
Given the frozen backbone and the previous processing steps, this context integration is grounded in semantic meaning, section type, and aligned across languages. \\
For each document $x_d$, a new sequence $E''_{\text{token}_{d}}$ is obtained from $E'_{\text{token}_{d}}$, such a:
\begin{equation}
\operatorname{Head}(E'_{\text{token}_{d}}) = E''_{\text{token}_{d,l}} = \{\vb*{e''}_{\text{token}_{d,i}} \mid i=1\dots n \}.
\end{equation}
This ensures that each token's representation is both comprehensive and contextually aware, reflecting the document as a whole. \\
Finally, empirical tests have demonstrated that incorporating a skip connection layer between the output of the frozen backbone embeddings $E_{\text{token}{d,l}}$ for all $l \in L_\text{section}$ and the embeddings produced by the head $E''_{\text{token}_{d,l}}$ facilitated the training of the head. We hypothesize that this improvement arises from the head's ability to better adapt to the pretrained backbone by learning the difference between the generic domain and our specific domain.

\subsubsection{From embeddings tokens to vector representation}
\label{emb_to_vec}

Once the embedded tokens $\vb*{e}''_\text{token} \in E''_{\text{token}_{d}}$ are obtained for a document $x_d$, the final goal is to derive a single representation vector, denoted as $\vb*{e}_d$. To achieve this, a final pooling layer is applied. 

Our experiments indicate that using a weighted average pooling method yields better results than simple average pooling, which assigns equal importance to each token in the document sequence. This improvement is due to the fact that different sections within a document may vary significantly in length and relevance, making some tokens, such as those from a lengthy description, less significant than others, like those from a job title. To mitigate this issue, we set the pooling weights inversely proportional to the section length. Taking into account the skip connection, the final representation vector is computed as follows:
    \begin{equation}
    \vb*{e}_d = \sum_{l = 1}^{|L_\text{section}|} \sum_{j=1}^{k_l}
    \frac{\vb*{e}''_{\text{token}_{d,j+(\sum_{i=0}^{l}k_i)}} + \vb*{e}_{\text{token}_{d,l,j}}}
    {k_l . |L_\text{section}|}.
\end{equation}
This operation effectively performs pooling in two stages: first within each section, and then across all sections, resulting in a more balanced and representative final document representation vector.

\subsection{Training objective}

For structuring the latent space based on similarity between profiles and project proposals, a contrastive learning paradigm is adopted. Hence, a two-tower approach is used, with one tower for profiles and the other for project proposals. Finally, a contrastive loss function is applied for optimization. \\
Those kinds of loss leverage the relationship between positive and negative pairs of elements. During training, the model can receive input in the form of pairs corresponding to an anchor and a positive or an anchor and a negative element ; triplets corresponding to an anchor, a positive, and a negative element, or n-uplets with multiple positives and negatives elements.\\ Implementing the loss function for n-uplets allows for a more flexible and generalized approach better aligned with modern deep learning frameworks, and can also cover the simpler pair and triplet cases. \\
In the following sections, we will first explore the details of the contrastive approach in \ref{constrastive_approach}, then generalize our method to n-uplets in \ref{nuplets_section}. Next, section \ref{weaknegative_section} emphasizes the value of incorporating weak-negative interactions, followed by section \ref{adjancymatrix}, where we propose various loss functions tailored to our domain.

\subsubsection{The contrastive loss approach}
\label{constrastive_approach}

The training objectives in a contrastive setting aim to optimize the similarity or distance between document embeddings by maximizing alignment for positive relations and minimizing it for negative ones. This framework can be applied to both supervised and unsupervised contexts. \\
Among these methods, the triplet loss \cite{schroff2015facenet}, originally derived from the computer vision paradigm, operates on triplets of entities denoted as $(d, d^+, d^-)$. Here, $d$ serves as the anchor, $d^+$ is the positive example relative to the anchor, and $d^-$ is the negative example. Given a distance function $\mathcal{D}$ and the document embeddings $\vb*{e}_d$, $\vb*{e}_{d^+}$, and $\vb*{e}_{d^-}$, the triplet loss is formulated as:
\begin{equation}
\mathcal{L}_\text{triplet}(d, d^+, d^-) = \operatorname{max}\Big(\mathcal{D}(\vb*{e}_d, \vb*{e}_{d^+}) - \mathcal{D}(\vb*{e}_d, \vb*{e}_{d^-})+ m, 0\Big)
\end{equation}
where $m$ represents a specific margin. This loss function encourages the model to ensure that the positive example is closer to the anchor than the negative example by a margin of at least $m$, refining the embedding space.
Lately, a more commonly used loss function is the supervised contrastive InfoNCE loss \cite{oord2018representation}. For an entity $d$, this loss function can be generalized to work with multiple positive examples $D$ and negative examples $D'$. It is expressed as:

\begin{equation}
    \mathcal{L}_\text{InfoNCE}(d,D,D') = \eta
     \cdot \sum_{d' \in D}
        -\operatorname{log} \frac{\operatorname{exp}(\vb*{e}_d . \vb*{e}_{d'}/\tau)}{\sum\limits_{d'' \in D\cup D'}\operatorname{exp}(\vb*{e}_d . \vb*{e}_{d''}/\tau)}
\end{equation}
with a temperature parameter $\tau$ and $\eta = \frac{1}{|D|}$. \\
Both aforementioned losses can be generalized to handle n-uplets. It is constructed by considering the relationships between multiple elements, as detailed in the following sections.

\subsubsection{Generalizing to n-uplet}
\label{nuplets_section}
\begin{figure}
    \centering
    \includegraphics[width=\linewidth]{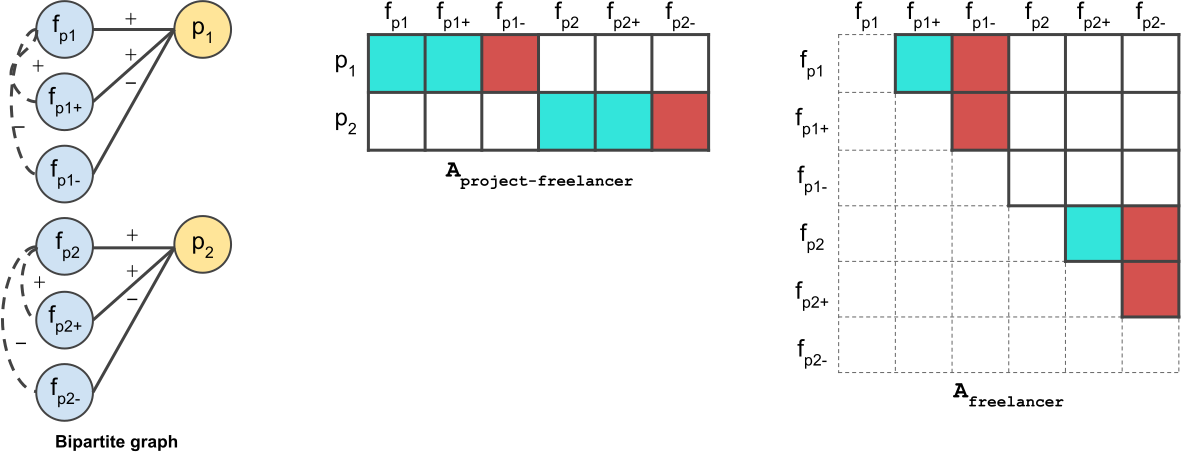}
    \caption{Bipartite graph and sub adjacency matrices of six triplets: - two freelancer triplets ($f_{b1}, f_{b1+}, f_{b1-}$), ($f_{b2}, f_{b2+}, f_{b2-}$), - four project-freelancer triplets ($p_1, f_{b1}, f_{b1-}$), ($p_1, f_{b1+}, f_{b1-}$), ($p_2, f_{b2}, f_{b2-}$), ($p_2, f_{b2+}, f_{b2-}$)}
    \label{fig:adjacency}
\end{figure}

The relationships between elements can be modeled as a bipartite graph, as shown in \textit{Fig.} \ref{fig:adjacency}. In this figure, the nodes represent the entities—freelancers denoted as $f_{p1}, f_{p1+}, f_{p1-}, f_{p2}, f_{p2+}, f_{p2-}$, and projects $p_1, p_2$—while the edges depict similarity relations based on historical data, such as skill matches (+) or contrasting (-). In this setting, the adjacency matrix $A$ correspond to a useful representation of the project proposal-profile relationship for tensor and matrix computation. To computationally filter or weight document pairs in the loss function, we represent a positive skill match with $1$, a negative match with $-1$, and an unknown relation with $0$. This matrix can be preprocessed based on skill set agreements in historical data, as described in Section \ref{approach}. \\
Furthermore, to help structure the latent space, a freelancer-to-freelancer skill similarity relation using a project as a pivot can be computed. Specifically, if two freelancers are both a positive skill match for a project, it indicates that they share a subset of skills that qualify them to undertake the project. Conversely, if one freelancer is a positive match while the other is a negative match, it suggests that the required skills are possessed by one but not the other. The model is then expected to learn these similarities and differences. \\ 
In this setting, when training a model at the batch level, a subset of projects $P \subset \mathbb{P}$ and the corresponding set of selected freelancers $F \subset \mathbb{F}$ can be sampled, and the relevant portion of the adjacency matrix can be extracted :
\begin{equation}
A_\text{project-freelancer} = A_{p \in P, f \in F},
\end{equation}
for computational purposes. \\
Using this submatrix, we can compute the freelancer similarity relation and, by extension, derive the transitive freelancer-to-freelancer adjacency matrix $A_{\text{freelancer}} \in \{-1, 0, 1\}^{|F| \times |F|}$ in the following manner :
\begin{align}
    A_{\text{freelancer}} = {{{A_{p \in P, f \in F}}}^T}^2 
    &. (f > f')_{f \in F, f' \in F} \notag\\ 
    &. [1 - {{{(A<0)_{p \in P, f \in F}}}^T}^2]
\end{align}
This operation retains only the upper triangular part ($(f > f'){f \in F, f' \in F}$) and ensures that two freelancers who are negatively related to a project do not become positively related ($1 - {{{(A<0){p \in P, f \in F}}}^T}^2$). \\
During our experiments, we did not resort to a project-to-project similarity relation. Such a relation is more challenging to determine and is not directly obtainable from our current settings. Indeed, a freelancer being capable of solving two different projects does not necessarily imply that the projects are similar, as they could require different sets of the freelancer's skills. In contrast, freelancer relationships can be directly derived and even supplemented from various data sources.

\subsubsection{Adding weak negatives}
\label{weaknegative_section}

While learning a latent space organization in a contrastive setting, negative examples are crucial as they serve as counterexamples, helping to refine both broad scenarios and edge cases. However, if we rely solely on the skill match relation in historical data, we are limited to addressing only edge cases. In practice, freelancers are contacted based on their perceived relevance to a project, and if negative feedback is received, it indicates that the project is not suitable for the freelancer at a more granular level (\textit{i.e.} on finer details, not visible at first glance). \\
To introduce more trivial counterexamples, which we refer to as weak negatives, we utilize a categorical feature of the profiles. Specifically, when creating a profile, a freelancer $f \in F$ must select a job category $l_f \in L_\text{profile}$. By defining an arbitrary negative similarity relation between two freelancers, $A[f,f'] = -1$, when they belong to different categories $l_f \neq l_{f'}$, we can incorporate weak negative examples since these freelancers are expected to have different skill sets. The transitive freelancer to freelancer adjacency then becomes:

\begin{equation}
\begin{split}
A_{\text{freelancer}} =& ({{{A_\text{project-freelancer}}^T}^2} + A_{f \in F, f' \in F}) \\
& \cdot (f > f')_{f \in F, f' \in F}.
\end{split}
\end{equation}

These negative examples are intended to assist in the initial training iterations, aid convergence, and improve the organization of the latent space at a coarse granularity level.

\subsubsection{Adjacency matrix based contrastive losses}
\label{adjancymatrix}

We can reformulate the different contrastive loss functions using the previously mentioned adjacency matrix within our framework. Our goal is to obtain a loss function, $\mathcal{L}(P,F,A)$, that takes a set of proposals $P$, a set of freelancers $F$, and an adjacency matrix $A$ as input. \\
First, we introduce a triplet loss that is calculated for all possible document triplets, based on the distances between two sets of entities, D and D'. Afterward, we filter the results using the adjacency matrix to retain only the valid triplets. This loss function is referred to as $\mathcal{L}_\text{A-triplets}$ and is computed as follows:

    \begin{equation}
\begin{split}
    \mathcal{L}_\text{A-triplets}( D, D', A) = 
    &\sum_{d \in D} \sum_{d' \in D'}  \sum_{d'' \in D'}
    \mathcal{L}_\text{triplet}(d, d', d'')
     \\
     &.\mathds{1}_{A_{d,d'}>0, A_{d,d''}<0} .
    \end{split}
\end{equation}

For tensor manipulation, the triplet loss $\mathcal{L}_\text{A-triplets}$ can be efficiently computed by first calculating the loss for all possible document triplets, producing a tensor with the shape $\mathbb{R}^{|D| \times |D'| \times |D'|}$. This tensor is then filtered using an element-wise product with a mask tensor. The mask tensor is derived from the relevant submatrix of the adjacency matrix between the relations $D$ and $D'$, and it retains only the valid triplets specifically, where the second dimension of the tensor represents the positive example relative to the anchor, and the third dimension represents the negative example relative to the anchor. \\
The final triplet loss used in our experimentation is computed by summing two A-triplets. In the first case, the set of project proposals $P \subset \mathbb{P}$ is considered as the anchor, and in the second case, the set of profiles $F \subset \mathbb{F}$ serves as the anchor. In both scenarios, the positive and negative elements are drawn from the set of profiles $F \subset \mathbb{F}$. The resulting loss is defined as the following : 
\begin{equation}
    \begin{split}
            \mathcal{L}_\text{dual A-triplets}(P, F, A) = 
            &\mathcal{L}_\text{A-triplets}(P, F, A_\text{project-freelancer}) \\
            &+ \mathcal{L}_\text{A-triplets}(F, F, A_\text{freelancer}).
    \end{split}
\end{equation}

In our case, we want the InfoNCE loss to be fully supervised, meaning we want to control exactly which document are negative in the loss. To do so, we can leverage the adjacency matrix:
\begin{equation}
    \scriptstyle%
    \mathcal{L}_\text{A-InfoNCE}(D,D',A) = \frac{1}{|A>0|} . \sum_{d \in X}\sum_{d' \in X'}
        -\operatorname{log} \left(f(d, d', A)\right),
\end{equation}
with 
\begin{equation}
    f(d, d', A) =  \frac
        {\operatorname{exp}(\vb*{e}_d . \vb*{e}_{d'}/\tau) . \mathds{1}_{A_{d,d'}>0}}
        {\sum_{d'' \in D\cup D'}\operatorname{exp}(\vb*{e}_d . \vb*{e}_{d''}/\tau) . \mathds{1}_{A_{d,d'}\neq 0}}.
\end{equation}
From this implementation, the tensor computation can be done in two part, first computing all dot products between all possible pairs then applying a masked softmax and filtering on positive pairs.
Then consider both projects $P\subset \mathbb{P}$ freelancers $F\subset \mathbb{F}$ pairs and freelancer $F\subset \mathbb{F}$ only pairs and adding the weak negatives in $A$ computed from categories, we get:
\begin{equation}
\begin{split}
    \scriptstyle%
    \mathcal{L}_\text{dual A-InfoNCE}(P,F,A) = 
    &\mathcal{L}_\text{A-InfoNCE}(P,F,A_\text{project-freelancer}) \\
    &+ 
    \mathcal{L}_\text{A-InfoNCE}(P,F,A_\text{freelancer} + A_{f \in F, f' \in F})
\end{split}
\end{equation}
We note that for the A-InfoNCE the cosine similarity is better suited than a distance function to reflect on semantic similarity, as it implies the maximization of a cartesian product.

%% file: 4_results.tex
\section{Experiment}

With our experiment, we want to evaluate our choices on how to leverage both pretrained models and our document data, while also comparing our approach to a reference model and models based on state-of-the-art approaches. The different model evolution tested here are to reflect on the impact of modifying the weights of pretrained models in the semantic organization and the language alignment of the latent space, but also the influence of our solutions to exploit document content and structure.

\subsection{Models}
\label{models_description}

In the following, we compare our approach presented in section \ref{approach} against one reference model and several models and architectures we tested while building our final approach:

\begin{itemize}
    \item The first model, used as reference, is PaLM 2, specifically its \textbf{Gecko} version\footnote{\emph{textembedding-gecko-001} from May 2023}. This is a closed commercial model from Google, based on a decoder-style transformer architecture with over 10 billion parameters. It is fifty to one hundred times bigger than the other models evaluated here.
    \item The second model, later referred to as \textbf{S-BERT}, is a multilingual model from the Sentence BERT\cite{sentence-bert} family which is finetuned on our data. This approach is fairly similar to what has been proposed with conSultantBERT \cite{lavi2021consultantbert}. However, we use a different backbone and training loss.
    \item The third model is also based on a Sentence BERT \cite{sentence-bert}, but applied per section of text. This is followed by an average pooling of the resulting section embedding vectors to produce a single vector per document. For clarity, we refer to this model as \textbf{Section-S-BERT}.
\end{itemize}
Concerning the first two models (Gecko and S-BERT), both take project proposals and profiles as plain text inputs. To obtain only one piece of text, all sections are concatenated, while Section-S-BERT and our proposed method process the sections separately, without concatenation. S-BERT and Section-S-BERT are trained using the same multilingual SentenceBERT~\cite{sentence-bert} as backbone\footnote{https://huggingface.co/sentence-transformers/distiluse-base-multilingual-cased-v1} which is aligned through distillation~\cite{sentence-bert-multilingual}. Whereas, our approach leverages a multilingual E5~\cite{e5,e5-multilingual} as backbone\footnote{https://huggingface.co/intfloat/multilingual-e5-small}. This choice is significant because, unlike SentenceBERT, the E5 model does not project the final layer, which is crucial for our method, as discussed in section \ref{section_level}. \\
Some models were evaluated out-of-the-box in a zero-shot manner, while others were trained on our dataset, as described in the following sections.

\subsection{Dataset}

To avoid data contamination, the data used for the following experiments and testing were split in a temporal manner. Specifically, the test set consists of ten weeks of past interactions starting from March 2023, comprising approximately 61k interactions, 4k project proposals, and 32k freelancer profiles. For the training and validation data, which were split in an 80/20 ratio, all interactions that occurred prior to the aforementioned date were considered. Furthermore, all profiles were recomputed to reflect their state at the time of the interactions. \\
Concerning the freelancer profiles, various declared information are used as sections, including a free-text job title and description, a job family and category selected from our taxonomy, and a set of skills, which may be provided in free text or selected from those previously defined by other freelancers. For the project proposals, the retained information includes the mission title, its associated job title and description as free-text, a job family and category from our taxonomy, as well as mandatory and bonus skills chosen from the list of freelancer-declared skills, as illustrated in \ref{fig:methodillustration}.

We consider empty sections, referring to sections not filled by the freelancers or the companies, as sequences containing only the tokens [CLS] and [END]. This data modeling approach offers two key benefits: (1) it enables the use of static computational graphs for optimization, and (2) it provides the models with explicit information about the emptiness of the section.

\subsection{Training setting}

In this experiment, Gecko and Section-S-SBERT were evaluated in a zero-shot setting, while all other models were trained on our dataset, either with full retraining or with a frozen backbone. Specifically, we evaluated both S-BERT and Section-S-BERT with fully retrained backbones, starting from their initial pretrained states. Additionally, we trained a Section-S-BERT with a projection layer and our final proposed approach, both with their pretrained backbone weights frozen. In the case of Section-S-BERT with a projection layer, it should be noted that only the projection layer was trained. \\
Regarding the loss functions, almost all models were optimized using a triplet loss \cite{sentence-bert}. Our approach, as well as the Section-S-SBERT with a projection layer, were trained using the InfoNCE loss \cite{oord2018representation}. The trainings using the triplet loss were conducted over ten epochs with a batch size of two, \textit{i.e.}, two project proposals per batch. Based on these project proposals, one negative and two positive freelancers were sampled, enabling the creation of two freelancer triplets per batch, in addition to the four project-freelancer triplets. \\
For the trainings based on the InfoNCE loss, the models were trained for two epochs with one project proposal per batch. The same freelancer sampling strategy as used with the triplet loss was employed, but with the addition of thirty weak negative freelancers.

\subsection{Evaluation metrics}

\setlength{\tabcolsep}{8pt}
\begin{table*}[]
    \centering
    \begin{tabular}{llccccccccc}
    \toprule
        \multirow{2}{*}{Model} & \multirow{2}{*}{Loss} &
         \multicolumn{2}{c}{valid score recall} &
        \multicolumn{2}{c}{Category overlap} & \multicolumn{2}{c}{skills overlap} &
        \multicolumn{2}{c}{retieved@100} \\
         & & single & all & @10 & @100 & @10 & @100 & \% + & \% - \\ 
    \midrule
        $\diamondsuit$ Gecko & no training &
        0.437 & 0.380 & 0.532 & 0.444 & 0.499 & 0.428 & 0.245 & 0.179 \\ 
        $\clubsuit$ S-BERT & Triplet &
        \textbf{0.477} & 0.461 & 0.028 & 0.027 & 0.042 & 0.042 & 0.082 & \textbf{0.059} \\ 
    \cmidrule(l){1-2}
        \multirow{2}{*}{$\clubsuit$ Section-S-BERT} & Triplet &
        0.458 & \textbf{0.510} & 0.028 & 0.028 & 0.042 & 0.041 & 0.082 & \textbf{0.059} \\
         & no training &
        0.432 & 0.385 & 0.774 & 0.709 & 0.538 & 0.479 & 0.289 & 0.222 \\
    \cmidrule(l){1-2}
        \multirow{2}{*}{$\clubsuit$}
        Frozen Section-S-BERT & Triplet &
        0.451 & 0.413 & 0.610 & 0.535 & 0.413 & 0.369 & 0.195 & 0.160 \\
        $\quad$+ projection & InfoNCE &
        0.452 & 0.425 & 0.666 & 0.602 & 0.454 & 0.413 & 0.220 & 0.186 \\
    \cmidrule(l){1-2}
        $\spadesuit$ \textbf{ours} & InfoNCE &
        0.431 & 0.398 & \textbf{0.821} & \textbf{0.754} & \textbf{0.613} & \textbf{0.539} & \textbf{0.356} & 0.281 \\
    \bottomrule
    \end{tabular}
    \caption{Table containing the results of models with the training setting evaluated on the test data described. Best metrics values are set in bold. $\diamondsuit$: PaLM2, $\clubsuit$: multilingual sentence BERT, $\spadesuit$: multilingual E5 }
    \label{tab:retrieval_models_results}
\end{table*}

Each of the aforementioned models aims at encoding projects and freelancers for retrieval within the proposal step of a recommender system. However, evaluating them in the same manner as we would evaluate the full pipeline, using ranking or retrieval evaluation metrics, is not perfectly suited. Moreover, the proposal-profile relevancy, in the context of document retrieval, is only partially available in our dataset, as the relevancy of all possible proposal-profile pairs remains unknown. Finally, evaluating solely based on past interactions may not provide a complete picture of the intended performance of a retrieval system. Indeed, historical data is biased by previous processes and lacks information about preposterous project-freelancer pairs. \\
To better reflect the future usage of the models, we chose to evaluate them in three different settings:
\begin{itemize}
    \item \textbf{Supervised}, using only historical interactions and the interacted profiles,
    \item \textbf{Unsupervised}, by simulating a retrieval with k-nearest-neighbor (k-NN) using all test profiles,
    \item \textbf{Weakly supervised}, using the simulated retrieval enriched with the historical interactions.
\end{itemize}
The cosine similarity or Euclidean distance was used to compute the predicted proposal-profile score when the training loss was InfoNCE or the triplet loss, respectively. Both were converted to scores or distances according to the requirements of the metrics and the k-NN algorithm. \\
Considering the \textbf{supervised} experiments, we tested the recall metric using score comparison, denoted as $s(\cdot)$, to determine true and false positives. This metric, which we will refer later to as "\textit{valid score recall}," is computed as follows:
\begin{equation}
\begin{split}
        &\textit{recall}_\textit{single}(p, F_+, F_-) = \\
        &\sum_{f_+ \in F_+} \frac{
        \mathds{1} \left[ \operatorname{s}(\vb*{e}_p, \vb*{e}_{f_+})
        > \operatorname{s}(\vb*{e}_p, \vb*{e}_{f_-}), \forall f_- \in F_-
        \right]
    }{|F_+|},
\end{split}
\end{equation}

where $p$ is a project proposal, $\vb*{e}_p$ corresponds to its encoding, $f_+ \in F_+$ the relative positive freelancers, $f_- \in F_-$ represents its negative freelancers, and $\vb*{e}_{f_+}$, $\vb*{e}_{f_-}$ are their respective embedding vectors. \\
In this setting, we also compare the average of positive score against the average score of negatives, with the following :
\begin{equation}
\begin{split}
    &\textit{recall}_\textit{all}(p, F_+, F_-) = \\
    &\mathds{1} \left[ 
    \left[ \sum\limits_{f_+ \in F_+}
        \frac{\operatorname{s}(\vb*{e}_p, \vb*{e}_{f_+})}{|F_+|}
    \right]
    > \left[ \sum\limits_{f_- \in F_-}
        \frac{\operatorname{s}(\vb*{e}_p, \vb*{e}_{f_-})}{|F_-|}
    \right] \right].
\end{split}
\end{equation}
Inverting the comparison and using a distance function allows for the computation of the metric when the triplet loss was used. \\
For the \textbf{unsupervised} setting, a retrieval is first simulated using a k-NN algorithm for each project on all freelancers in the test set. Then, the proportion of freelancers who have declared the same category as the one in the project proposal is computed. \\
We introduce the \textit{A-overlap} metric to quantify lexical similarity between freelancers and project proposals using exact lexical matching on a predefined set of terms:
\begin{equation} \textit{A-overlap}(A; p, F) = \sum_{f \in F}\frac{|A_{p} \cap A_{f}|}{|F| \cdot |A_{p}|},
\end{equation}
where $A$ denotes either the set of skills $S$ or of categories $C$. The overlap score is calculated for each freelancer, and the results are averaged across all retrieved freelancers $F$.
\\
Finally, for the \textbf{weakly supervised} setting, both the simulated retrieval and the historical interaction data are combined. This allows us to compute the proportion of retrieved positives, such as : 
\begin{equation}
    \textit{retrieved-positves}(p,F,F+) = \frac{|F \cap F_+|}{|F_+|},
\end{equation}
and retrieved negatives : 
\begin{equation}
    \textit{retrieved-negatives}(p,F,F-) = \frac{|F \cap F_-|}{|F_-|},
\end{equation}
among the retrieved freelancers $F$.

\subsection{Results}

\begin{figure*}[]
\centering
{\includegraphics[width=0.8\textwidth]{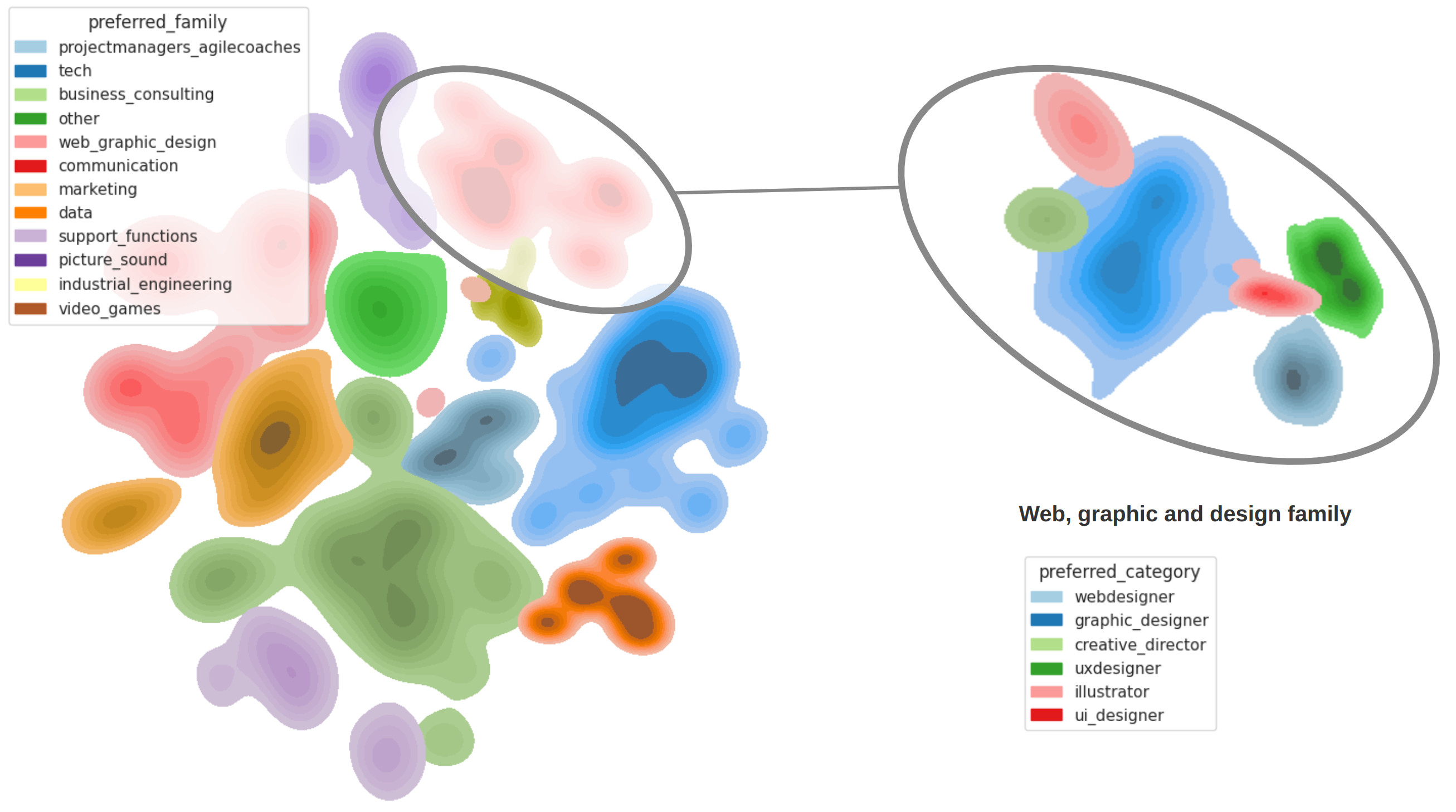}}%
    \caption{KDE Density plots of obtained profile embeddings projected in two dimensions using T-SNE. Color encodes the family or job category associated with the profile embeddings. On the right we zoom in the web, graphic and design family. }
\label{freelancers_latent}
\end{figure*}

The following results, reported in table \ref{tab:retrieval_models_results},  consist of the metrics defined previously, computed across all models described in Section \ref{models_description}. Each metric represents the average value computed over all project proposals in the test dataset. \\
Gecko and Section-S-BERT, corresponding to the first and fourth lines of Table \ref{tab:retrieval_models_results} are evaluated in a zero-shot manner, and provide strong results for the unsupervised metrics related to category and skill overlap. These results reflect the effectiveness of the semantic-aware pretraining and the models' ability to organize documents in the latent space based on their content. However, due to the lack of fine-tuning on our dataset, the supervised valid score recall metric reveals these models' inability to accurately structure the latent space according to our data domain and historical interactions. \\
Adapting the models to our domain through full retraining yields improved results, as shown in the second and third rows of the table. However, the low values obtained for the weakly-supervised and unsupervised metrics suggest a degradation in the semantic space organization established during the pretraining phase. \\
The results reported in the last three rows of the table, corresponding to models with frozen backbones, appear to offer the best of both worlds. These models successfully retain the benefits of the semantic pretraining of the backbones while adapting to our domain. This is evidenced by the improved supervised metrics compared to the fully retrained versions, without significantly degrading the unsupervised metrics. \\
Adding synthetic weak negatives in the training improves the unsupervised metrics, as reported in the last two rows of the table. We hypothesize that this improvement is due to the backbones being adapted to both edge cases and broader scenarios using more extensive data. In contrast, the other models adapt to our domain by focusing exclusively on edge cases during training. \\
Finally, our latest approach appears to be the most effective in leveraging content. Not only does it avoid degrading the unsupervised metrics, but it also achieves the best overall results in this regard. It seems to adapt the backbone to edge cases based on historical data while simultaneously adapting it to our domain, being 50 times smaller than PALM 2. It is important to note that our approach enables the retrieval of more positives than negatives in the historical data, although the number of negatives remains substantial. This should be carefully considered when designing the entire pipeline to ensure proper filtering of these freelancers. \\
In addition to the aforementioned quantitative metrics, we present the freelancers' latent space organization for our trained approach in \textit{Fig.} \ref{freelancers_latent}. A two-dimensional projection of the freelancers' density is illustrated both by job family in the left of the Figure, and by category in its right, specifically for the web, graphic and design job family. When focusing on the density distributions by family, denser spots can be observed. Indeed, some families exhibit a single dense spot, while others display multiple dense areas. \\
When exploring the latent space representing the job categories, it can be observed that the previously mentioned denser spots correspond to regions of space reserved for categories within the families, even though some categories may have multiple dense spots. We hypothesize that this may be due to the presence of sub-categories and different typologies of freelancers. \\
This latent space can be leveraged not only for retrieval, but also to gain a deeper understanding of our market and freelancers. In particular, it could be used to drive the evolution of our job category taxonomy. Overall, our model appears to effectively organize freelancers based on semantic relationships. \\
In conclusion of these experiments, we report metrics based on the language alignment aspect of our approach, as shown in Table \ref{tab:retrieval_languages_results}. These results highlight the model's ability to maintain pretrained language alignment while simultaneously adapting the backbone to our domain.
This result is particularly notable given that our dataset contains approximately ten times more French profiles than profiles in other languages. Hence, achieving effective language alignment solely based on this data is challenging without the use of pretrained backbones, highlighting the good performances of our approach.

\setlength{\tabcolsep}{2pt}
\begin{table}[]
    \centering\footnotesize
    \begin{tabular}{ccccccccc}
    \toprule
        \multirow{2}{*}{language} & project & score &
        \multicolumn{2}{c}{
            cat-overlap
        } & \multicolumn{2}{c}{
            skills-overlap
        } &
        \multicolumn{2}{c}{retieved@100} \\
         & support & $recall_\textit{all}$ & @10 & @100 & @10 & @100 & \% + & \% - \\
    \midrule
        French & 3169 & 0.402 & 0.832 & 0.770 & 0.618 & 0.551 & 0.337 & 0.263 \\
        
        Spanish & 341 & 0.351 & 0.768 & 0.670 & 0.662 & 0.551 & 0.439 & 0.346 \\

        English & 209 & 0.369 & 0.763 & 0.699 & 0.455 & 0.400 & 0.308 & 0.222 \\
        
        German & 185 & 0.454 & 0.793 & 0.695 & 0.601 & 0.460 & 0.560 & 0.489 \\
        
        Dutch & 7 & 0.286 & 0.686 & 0.724 & 0.669 & 0.578 & 0.506 & 0.333 \\
    \bottomrule 
    \end{tabular}
    \caption{Metrics results of our approach on different languages available in our test data. The score recall correspond to the "valid score recall: all"}
    \label{tab:retrieval_languages_results}
\end{table}

%% file: 5_industrial.tex
\section{Deployment in production}

As outlined in Section \ref{sec:intro}, a legacy recommender system with one filtering phase and one ranking phase was already in place. This section details the integration of the newly developed retriever within the existing infrastructure and examines the impact of its deployment in our production environment.

\subsection{Infrastructure}

As illustrated in \textit{Fig.} \ref{fig:methodillustration}, we implemented a conventional architecture for deploying our neural retriever. The embeddings for all freelancer profiles are precomputed and stored in a vector database. For this purpose, we selected Qdrant\footnote{https://qdrant.tech/}, due to its high performance and advanced filtering capabilities, including geographical filtering. The vector database is updated daily to incorporate new and updated profiles.

At inference time, upon receiving a new project proposal, its embedding is computed on the fly. Then, we use that embedding to perform an approximate nearest neighbors search taking into account hard filters to retrieve a set of candidates for the projects. These candidates are subsequently passed to the legacy ranking system for final evaluation.

\subsection{A/B test}

To ensure that the new retrieval phase did not negatively impact conversion rates, an A/B test was conducted over the course of November 2023. As anticipated, there was a significant improvement in response times, with the 95th percentile latency decreasing from tens of seconds (and occasionally exceeding a minute) to a maximum of 3 seconds. More unexpectedly, a 5.63\% improvement in conversion for effective matches (\textit{i.e.}, cases where both the freelancer and client confirm the match) was observed. We hypothesize that the new retriever, with its more advanced technology, effectively eliminates suboptimal candidates that might have been selected by the legacy ranker.

%% file: 6_futur.tex
\section{Conclusion}

This research aimed to develop two encoder models that generate embeddings for freelancer profiles and project proposals, enabling efficient vector-based retrieval. The models were designed to handle multilingual data, exploit document structure and content, and scale effectively as part of a candidate proposal system.\\
We proposed and validated an architecture and training approach, which outperformed other methods we tested. When deployed in production, it improved latency without any matching performance loss, leading even to an improved conversion rate.\\
Future work could explore improving data preparation for contrastive learning, such as creating better positive and negative pairs. Additionally, using continuous values in the adjacency matrix, rather than binary edges, could enhance learning semantic similarity, particularly in synthetic relationships between freelancers. Combining lexical and semantic retrieval methods, such as integrating BM25 scores, also presents a potential area for improvement. \\
While our models are effective for recommendations, they are less suited for search tasks where the input are just short queries instead of complete project proposals. A possible direction for future research could involve a three-tower architecture, with one tower dedicated to handling search queries. \\
Finally, an important area of research within human resources technology involves studying biases induced by such models. That topic is deeply important to Malt and for the AI Act. That is why we plan on investigating and mitigating potential biases in our retriever models. This would ensure fairness and transparency in the recommendations, while exploring techniques to identify and correct any unintended biases that may arise during training and deployment.